\documentclass[12pt]{article}
\usepackage{amsmath, amsthm, amssymb}
\usepackage{graphicx}
\usepackage{enumerate}
\usepackage{multirow}
\usepackage[sort]{natbib}
\usepackage{shortcuts}
\usepackage{booktabs}
\usepackage[capitalise]{cleveref}
\usepackage{suffix}
\usepackage{enumitem}
\usepackage{bm}

\newcommand{\A}{\mathcal A}
\newcommand{\sumA}{\sum_{a\in\A}}

\WithSuffix\newcommand\sumA*[1]{\sum_{#1\in\A}}

\def\blind{0}

\if\blind1
\usepackage{xcolor}

\fi
\if\blind0

\fi

\begin{document}

\def\spacingset#1{\renewcommand{\baselinestretch}%
{#1}\small\normalsize} \spacingset{1}

\newcommand{\mytitle}{Rejoinder: New Objectives for\\Policy Learning}
  \title{\bf \mytitle}
  \if1\blind\author{}\fi
  \if0\blind\author{Nathan Kallus\thanks{School of Operations Research and Information Engineering and Cornell Tech, Cornell University}}\fi
  \date{}
  \maketitle

\bigskip
\spacingset{1.5} %

I would like thank the discussants,
Oliver Dukes and Stijn Vansteelandt (DV),
Sijia Li, Xiudi Li, Alex Luedtkeand (LLL),
and Muxuan Liang and Yingqi Zhao (LZ),
for a very thoughtful discussion both of my 
contribution \citep{kallus} and of \citet{mo2020learning}. I similarly thank 
the editors for putting together this 
exciting special issue and for curating a 
timely discussion on new objectives for policy learning.
I found the juxtaposition between the two papers
particularly apt: while my paper tries to induce an optimal 
covariate shift based on the 
premise of invariance, \citet{mo2020learning} 
try to be robust to an undesirable covariate shift for fear of 
variations.
While one optimistically alters the training population, the other 
pessimistically considers the worst-possible testing population.

In the following I review some discussant comments that stood 
out to me as particularly keenly perceptive and offer some 
reflections. I review DV's very observant discussion of the imperative to retarget, especially in analyses of electronic health records. I try to flesh out some of the connections alluded to by DV and LZ between retargeted policy learning and predictive modeling of treatment outcomes and effects. At LLL's apt urging, I consider whether accounting for curvature (or gap or margin) in policy value might help choose better retargeting weights. Finally, I concur with LLL's comments on \citet{mo2020learning} on the use of data from the testing distribution and offer a few additional thoughts.

\section{The Imperative to Retarget}

DV observantly outline some important practice-driven rationales for 
retargeting the policy learning objective.
First, targeting a particular population such as the training-data-generating one might be too ambitious, as would be evidenced in high variance value estimates when overlap is weak.
Second, DV provide an expert portrayal of the unspecialness of the training covariate distribution as a 
target population in analyses of 
electronic health records, so that we might as well 
choose an easier target population.

Third, DV point to an important rationale I had not
mentioned or thought of: when the observed distribution of 
treatment choices $A$
represents current practice, as in the example of electronic
health records, retargeting the population using my proposal
also focuses the learning on the covariate region where decision
makers, \eg, doctors, were historically 
unsure about how to treat and therefore 
vary most, \ie, covariates $X$ having $\max_{a\in\mathcal A}\phi(a\mid X)$ far from $1$. 
If doctors consistently and overwhelmingly choose a
certain treatment in certain cases (the above is close to $1$), most likely it is a good 
treatment. This is of course not guaranteed, but either way
convincing them to act otherwise in these cases may be hopeless.
Emphasizing value of focusing on improving decisions where previously there was uncertainty or equipoise is an excellent point; fortunately this is also exactly where counterfactual evaluation is most feasible.

DV conclude from this that we should use the variant of my proposed 
retargeting weights that assumes homoskedasticity, \ie, $w_0(x)\propto\prns{\sumA\frac1{\phi(a\mid x)}+\frac{m}2-1}^{-1}$. I agree that focusing on these weights is most practical in applications as it avoids having to additionally estimate the conditional variances, which may have only little to no additional benefit, and I should have made this clearer. Indeed, these are the weights I used in the job-counseling case study I considered in my paper, without attempting to estimate the conditional variances even though they are conceivably non-constant in the data, which comes from a real trial.

Conceptually, nonetheless, one could argue that residual outcome variance does also account for which are the ``subjects about which clinicians are sufficiently uncertain which policy is best,'' as DV write. Higher residual variance means any decision maker faced with finite data, doctors included, would necessarily encounter more uncertainty. Although to the extent that the data -- and the choice of covariates -- did not actually exist before they were constructed by an analyst processing electronic health records into a covariate-action-outcome dataset, I concede that this may be a somewhat stylized interpretation.

\section{New Objectives for Causal Prediction}

DV note that for decision support, a prediction of the outcome of an intervention might be more helpful than a policy recommendation in some settings, and they astutely point out that retargeting may be useful for this task as well. LZ also point out connections between retargeting weights and efficient estimation of effect modifiers.
DV propose an interesting overlap-weighted least-squares regression for the mean outcome of each treatment. I conjectured this might be motivated either by trying to estimate a best linear fit or a correct linear conditional mean, but either case seems to lead to an estimating objective different from DV's. Here we focus on the binary-action setting, $\mathcal A=\{-,+\}$.
We focus on linear models for simplicity but this applies more generally.

\paragraph{Estimating the best fit}
Suppose we seek the best fit for the treated outcome $Y(+)$ as a linear function of $X$ on the $w$-weighted population, $\beta\in\argmin_\beta\E[w(X)(Y(+)-\beta^\top X)^2]$, or equivalently, $\E[w(X)(Y(+)-\beta^\top X)X]=0$.
One example is of course $w=\bm 1$.
The latter estimating equation is expressed in terms of the unavailable complete data. Using \citet{tsiatis2007semiparametric} we can convert it into the efficient estimating equation using observed data:
\begin{equation}\label{eq:bestfit}\E[w(X)(\psi_+-\beta^\top X)X]=0,\end{equation} where $\psi_+=\mu(+\mid X)+\frac{\indic{A=+}}{\phi(+\mid X)}\prns{Y-\mu(+\mid X)}$ using the same definitions for the mean outcome $\mu$ and propensity $\phi$ as in my paper. 
Thus, an efficient estimate may be given by a $w$-weighted least-squares regression of the doubly-robust pseudo-outcome (\eg, using cross-fitted nuisance estimates with certain slow-rate conditions) on $X$. 
We can of course also focus on any other function class, for example $\beta^TZ$ for some coarsening $Z=\phi(X)$, which might pick out only a subset of the covariates.

\paragraph{Estimating the mean outcome}
The above is \emph{different} from assuming $\E[Y(+)\mid X]=\beta^\top X$.
In this case, the efficient estimating equation for $\beta$ is 
\begin{equation}\label{eq:linear}\Eb{\frac{\indic{A=+}}{\sigma^2(+\mid X)}(Y-\beta^\top X)X}=0,\end{equation}
\ie, an efficient estimate is given by running precision-weighted least squares regression of $Y$ on $X$ only on the treated data ($A=+$) and \emph{ignoring} the untreated data. This makes sense because both the training and testing covariate distributions are ancillary to a well-specified parametric regression. Note this can be rewritten as
$\E[\frac{\phi(+\mid X)}{\sigma^2(+\mid X)}(\psi_+-\beta^\top X)X]=0$, 
which is regressing the doubly-robust pseudo-outcome on $X$ on the $\frac{\phi(+\mid X)}{\sigma^2(+\mid X)}$-weighted population. So, roughly, compared to \cref{eq:bestfit}, assuming linear is well-specified allows us to target this easier population.
In practice, to avoid having to estimate $\sigma^2(+\mid X)$, we can either ignore the precision weight and just use ordinary least squares or use iteratively reweighted least squares (IRLS) with inverse squared residual weights.
If we instead assume $\E[Y(+)\mid Z]=\beta^\top Z$ for some $Z=\phi(X)$ of interest, we still have that precision-weighted least squares of $Y$ on $Z$ only on the treated data suffices.

\paragraph{DV's proposal} DV's proposal for a linear model
 can be read as the estimating equation 
\begin{equation}\label{eq:dv}\E[\indic{A=+}{\phi(-\mid X)}(Y-\beta^\top X)X]=0,\end{equation}
\ie, $\phi(-\mid X)$-weighted least squares regression of $Y$ on $X$ only on the treated data ($A=+$), which DV point out is consistent for the best linear fit for the treated outcome on the ${\phi(-\mid X)\phi(+\mid X)}$-weighted population. But it is not generally efficient for this estimand -- that would be \cref{eq:bestfit} instead. 
The proposal for more general $Z=\phi(X)$ provided by DV in their appendix involves additional nuisances but still appears to be generally inefficient.\footnote{The derivation is also unclear. The conditional restriction considered in the appendix is equivalent to $\E[Y(+)\mid X]=g_1(Z)$, \ie, the conditional mean of treated outcome given $X$ is a function of only $Z$. 
The final loss function, appearing in their last display equation, also appears to me to be in error.}
Moreover, if linear is not well-specified, the best linear fit depends on the covariate distribution and it is not clear why this particular population should be the target of interest. If variance and simplicity were the aim despite uninterpretability, it might make more sense to target the $\phi(+\mid X)$-weighted population, where we would simply regress $Y$ on $X$ in the treated data.
If we do assume linear \emph{is} well-specified, $\E[Y(+)\mid X]=\beta^\top X$, then \cref{eq:dv} is in fact equal to \cref{eq:bestfit} with $Z=X$ and $w(X)={\phi(-\mid X)}\phi(+\mid X)$. However, if we assume this, then we should instead just be using \cref{eq:linear} to be efficient. And, using \cref{eq:linear} is not any harder and, if we ignore the precision weights or use IRLS, requires no nuisances. Compared to this, \cref{eq:dv} has an extraneous weighting by $\phi(-\mid X)$.

\paragraph{An Interpretability-Variance Tradeoff}

DV mention an ``interpretability-variance tradeoff'' that arises in retargeted policy learning. The same tradeoff arises in causal prediction, as the above shows. If a linear (or some other parametric) model is well-specified, we can train on a more convenient population, such as simply the population where we get labeled observations, but if it is not then the result of doing so may be hard to interpret. The same tradeoff even occurs in classic linear regression without missingness or causal structure: if linear is well-specified then precision-weighted least squares is efficient, but if it is not then this yields a best linear fit on some hard-to-interpret population while ordinary least squares yields the best linear fit on the data-generating distribution. Without missingness, however, the differences are usually small and model validation is easy to do anyway. Therefore, the tradeoff is arguably much more central to causal or missingness settings, but it is a tradeoff whether we are doing policy learning or predictive modeling.

\paragraph{Treatment effect}
We can repeat this for $Y(+)-Y(-)$. If we want its best fit as a function of $X$ on the $w$-weighted population, the efficient estimating equation is
$\E[w(X)(\psi_+-\psi_--\beta^TX)X]=0$ where $\psi_-$ is the analog to $\psi_+$, corresponding to least-squares regression of the pseudo-outcome for the effect on $X$. \citet{kennedy2020optimal} provide strong error bounds for such an approach in very general and nonlinear settings. \citet{nie2017quasi} can also be viewed as an overlap-weighted version of such an approach. If we assume a linear effect function is well-specified, $\E[Y(+)-Y(-)\mid X]=\beta^\top X$, efficient estimation procedures for $\beta$ are reviewed in \citet[Section 3.2]{vansteelandt2014structural}.
LZ point out that \citet{liang2020semiparametric} study the efficient estimation for $\beta$ when assuming $\E[Y(+)-Y(-)\mid Z]=g(\beta^\top Z)$ for an unknown $g$ (where $\beta$ may even be matrix), and interestingly the same retargeting weights as in either \citet{crump2009dealing} or my paper show up in the efficient score.

\section{Accounting for Curvature}

LLL astutely point out the importance of considering the \emph{curvature} of the policy learning objective, in \emph{addition} to its estimability. The curvature, essentially, is how much we are penalized for deviating slightly from optimality. In what might at first appear paradoxical to the statistical learning neophyte (as it did to me at first), high curvature in a noisy loss function is actually good for decision making. On the balance between higher penalties for errors and a higher signal-to-noise ratio, in the end higher curvature leads to smaller regret. \Eg, in multi-arm bandits, having nearly-optimal arms is actually bad.
The key is to consider how fast we learn the optimal choice itself.

I first want to point out that the notion of curvature does appear in my analysis in Section 5.3, albeit in an argument relying on a finite policy space rather than a twice-continuously-differentiable value function as LLL use. My aim in Section 5.3 is to highlight that retargeting leads us to ``learn the underlying optimal policy parameters faster, and can therefore generalize better to any population, including the original one.'' In Lemma 5.4, $\gamma_n(w)=\max_{\pi\in\Pi_0} V_n(\pi;w)-\max_{\pi\in\Pi_0\backslash\Pi_{0,n}^*(w)} V_n(\pi;w)$ is the relevant curvature quantity for a finite set of choices: it is the value gap between any best and any second-best policy, \ie, how much we are penalized for deviating slightly from optimality.\footnote{Unfortunately, a typo in the Journal version replaced the two $\max$'s in its definition with two $\min$'s. In private correspondence, LLL, who pointed out the typo, explain this typo is the cause for their mistaking this quantity for the value diameter of the policy space (display equation above their Equation (5)) rather than the optimality gap.} The Lemma shows that, when optimizing the efficiently-estimated $w$-retargeted objective, the probability of choosing a suboptimal policy is crucially controlled by $\Omega^{1/2}(w,\rho)/\gamma_n(w)$.
When we choose the optimal policy our regret is zero on \emph{any} test population, and when we choose wrong one it is bounded; so, on average it is very small, and this can be applied to \emph{any} test population.
This quantity is the direct analog to LLL's $\Omega^{1/2}(w,\rho)/V''(\theta^\sharp;w)$ appearing in their Section 2.3, where $\theta^\sharp$ refers to the optimal policy parameter.
Indeed, in the smooth-value-function case, the second derivative quantifies the penalty we incur as we deviate slightly from optimality. I thank LLL for pointing out the curvature argument can be carried out in the smooth case, similarly to my finite case.

In the subsequent discussion after the Lemma, I argue that there is no a priori reason to believe $\gamma_n(w)$ is smaller or larger for the original population ($w=\bm 1$) or the optimally-retargeted one ($w=w_0$), and therefore we should focus on making $\Omega(w,\rho)$ small.
While I still believe the first statement is true (we should have no prior belief on the curvature/gap/margin under different $w$), I am intrigued by LLL's very appealing proposition: \emph{estimate} the curvature from data and incorporate it into the population-choosing objective. While I think this an innovative and very exciting direction forward, I am still somewhat unconvinced that we currently have a good way to adapt to different curvatures.

I want to first reemphasize that there is nothing special about the original population ($w=\bm 1$) with regard to curvature: it may better, it may be worse. The flat-value-surface example that LLL pose at the end of their Section 2.2 can happen for $w=w_0$ just as much as it can happen for $w=\bm 1$. In fact, rather than view the effect of curvature via LLL's Equation (3) focusing on the ratio of regrets in two populations, I believe the conclusion of their Taylor expansion in the display equation preceding this Equation might better be understood as two separate statements.
First, the regret of $\theta$ in \emph{any} test population \emph{at all} is controlled by the parameter error $\theta-\theta^\sharp$.
Second, if we learn $\theta$ by optimizing the $w$-retargeted objective, the error $\theta-\theta^\sharp$ is controlled by the $w$-retargeted regret.\footnote{Both statements depend on the relevant smoothness, of course. More generally and in multivariate settings, this can be phrased in terms of Lipschitz gradient (first statement) and strong convexity (second statement) of the value function. Alternatively, in the finite-policy-space case, we have the argument above using the probability of optimal choice.}
The important take-away is that, for the learning part, we need only care about the latter. Once we have a $\theta$ with small error, we can apply it well in \emph{any} population. From a theoretical point of view, barring any misspecification, we need not put any special emphasis on either the training or test populations in the learning stage, even if we know them.

Therefore, while I am enticed by the prospect of estimating and leveraging curvature information, I found LLL's restriction to $w_t=(1-t)w_0+t\bm 1$ in their Equation (4) confusing as it paints $t$ as trading off variance for curvature. Increasing $t$ may well deteriorate both. Since $w=\bm 1$ plays no special role, we might consider interpolating $w_t=(1-t)w_0+t w_1$ to other anchors $w_1$. But, we also know that choosing $w_1\propto\delta_\text{Dirac}(x-\theta^\sharp)$ (and $t=1$) would be optimal in LLL's Equation (4) so it would be farcical to choose another, even though such degenerate weighting will lead to very bad policy learning.
Going beyond simple examples where $V''(\theta^\sharp;w)$ exists (\eg, it might not for tree policies) and can be derived analytically and estimated (\eg, it cannot in moderate dimensions, for neural net policies, \etc) also appears to be very difficult.
It is also unclear how to interpret LLL's ``global curvature'' constraint: in fact, at the end of their Section 2.3, LLL note that it actually has the \emph{opposite} effect of constraining the true curvature.
One way to quantify the sharpness of the decision margin is in terms of the density near zero of the local action gap, $\Delta(X)=(\max_{a\in\mathcal A}\mu(a\mid X)-\max_{a'\in\mathcal A:\mu(a'\mid X)<\max_{a\in\mathcal A}\mu(a\mid X)}\mu(a'\mid X))_+$ 
\citep{mammen1999smooth,hu2020fast,luedtke2017faster} -- the less density near zero, the clearer the optimal action. And, this density changes as we change $w$. For binary actions, we have $\Delta(X)=M(X)=\max_{a\in\mathcal A}\mu(a\mid X)-\min_{a\in\mathcal A}\mu(a\mid X)$ (generally, however, $\Delta(X)<M(X)$) and thus LLL's proposal, which scales the proposed retargeted weights $w_0$ by $M(X)$, is minimizing $\Omega^{1/2}(w,\rho_0)/\E[w(X)\Delta(X)]$, up to scaling. This puts more weight where the mean-reward differences are large and the optimal action is obvious. But, this is not where statistical effort should be spent, in particular as suggested by LLL's curvature argument. Indeed, choosing $w$ so to obtain good $w$-weighted regret may be deluding oneself into believing a policy is performing well and is unhelpful on other test populations.
We might instead choose $w$ to minimize $\Omega^{1/2}(w,\rho_0)/\E[w(X)/\Delta(X)]$ in order to focus even more on the decision margin. I try this below but without great success.

\begin{table}[t!]
\centering\footnotesize
\begin{tabular}{ccccccc}
  \toprule
 Scenario & $\bm1$ & $\hat w_0$ & $\hat w^\ddagger=\hat w_0\hat \Delta$ & $\hat w_0\hat \Delta^2$ & $\hat w_0\hat \Delta^{-1}$ & $\hat w_0\hat \Delta^{-2}$ \\ 
  \midrule
1 & 0.005 (0.011) & 0.012 (0.020) & 0.011 (0.020) & 0.012 (0.021) & 0.037 (0.027) & 0.046 (0.024) \\ 
  3 & 0.033 (0.060) & 0.007 (0.018) & 0.022 (0.046) & 0.043 (0.067) & 0.008 (0.022) & 0.014 (0.031) \\ 
  4 & 0.015 (0.029) & 0.018 (0.036) & 0.021 (0.034) & 0.023 (0.030) & 0.054 (0.058) & 0.067 (0.057) \\ 
   \bottomrule
\end{tabular}
\caption{The mean regret (and standard deviation) in LLL's simulation study when using different (estimated) weights for policy learning (regret is on the unweighted population).}\label{table}
\end{table}

I am delighted that LLL included a very illuminating simulation study. I disagree, however, with their final conclusion that the study's results show that accounting for the curvature helps in practice: it appears to me there is no viable proposal that adapts to the different curvature settings.
Scenarios 1 and 2 by design have bad curvature specifically under $w=w_0$, while scenarios 3 and 4 are more neutral. In scenario 1 and 2 retargeting fails by design and uniform weights and LLL's two proposals do well, in scenario 3 the roles are reversed, and in scenario 4 all do equally well. There appears to be no method that adapts to each setting except for the oracle $w^\dagger$ method, which is of course not implementable.
I attempted to replicate LLL's results as well as try out the inverse weighting by $\Delta(X)$ proposed above. The results are shown in \cref{table}. Since LLL's code was not available, I did not replicate scenario 2 and the local curvature method as I focused on results requiring minimal changes to the existing replication code for my paper (the code to generate \cref{table} has been added to the replication repo at \texttt{https://github.com/CausalML/RetargetedPolicyLearning}). In each scenario, I consider taking the retargeted weights and multiplying them by $\hat\Delta^p=\abs{\hat\mu(+\mid X_i)-\hat\mu(-\mid X_i)}^p$ for $p\in\{-2,-1,1,2\}$, where $p=1$ corresponds to LLL's proposed $w^\ddagger$. Everything else is as in LLL, and all estimates are cross-fitted over two folds. While this replicates their results for retargeted and uniform weights, the results I obtain for their proposal are different than they report. The results for any $p\neq0$ appear to be bad. Among all methods, there appears no clear way to adapt to a curvature that benefits one population over another. The cases that favor the unweighted population are constructed to be so.

As stated at the onset, since there is no a priori reason to believe the curvature favors one population over another, it still appears wisest to me to focus on variance alone. I am no in no way ruling out that estimating and leveraging curvature information can help in selecting $w$. On the contrary, I think this is a very exciting and promising direction, but it may need further exploration. I thank LLL for pointing out this interesting idea and hope it inspires more researchers to consider it.

\section{A Regularity Problem}

One reason why it is difficult to directly consider the efficiency of learning the optimal policy parameters themselves is non-regularity. Therefore, we are left with inconclusive answers as in the above discussion. Generally this is insurmountable, but it is also what makes policy learning a very interesting problem.

When the true optimal policy parameters optimize a smooth convex surrogate loss, they may be regularly estimated, in which case we can more cleanly consider efficiency. Curiously, a reweighted empirical risk minimization of the surrogate loss does not appear to be efficient. Since being optimal for a surrogate loss induces a semiparametric model, efficient methods must incorporate a variety of moments, as shown by \citet{bennett2020efficient}, which does lead to some improvements in practice.

Going beyond surrogate losses, it is an interesting question in what other settings can we leverage regularity. For example, what can be gleaned under LLL's smoothness assumptions if we assume just a little bit more? Can this suggest optimal schemes in certain settings? This remains an interesting and important avenue for further research.

\section{Being Robust to the Test Population}

LLL point out that when samples from the test population are 
available, we should focus the method of \citet{mo2020learning} 
on ambiguity sets around the test population. They mention the idea
of using confidence intervals at level $1-\alpha$. This was
actually done in the context of distributionally robust optimization
using goodness-of-fit tests in \citet{bertsimas2018robust,bertsimas2018data}. However, such ambiguity sets may actually be too big, as pointed out by \citet{gupta2019near,lam2019recovering}. Ambiguity sets that make sure we bound the risk under $\mathbb P_X^*$ need not actually cover $\mathbb P_X^*$, and there is generally a ratio of square root of dimension between the size needed for the former and that needed for the latter.

An important question still is, even if the distributionally robust objective bounds the true unknown testing objective, when does this translate to robust policies. For example, if we were just adding a constant confidence term to the objective value of each policy, it would not change the final optimizer. As an example for distributionally robust optimization, using Wasserstein ambiguity sets to robustify a hinge loss recovers the original non-robust empirical risk minimizer \citep[Theorem 6.3]{esfahani2018data}. Since policy learning problems are essentially classification problems with missingness, the work of \citet{hu2018does} also appears particularly pertinent as it asks the question, ``does distributionally robust supervised learning give robust classifiers?'' Does distributionally robust policy learning? From the empirical results of \citet{mo2020learning} it appears it does sometimes, so it behooves us to understand when.

\section{Concluding Remarks}

The discussion by DV, LLL, and LZ is particularly exciting as it points to new and exciting avenues for research: new objectives for causal prediction, more careful considerations of the shape of the policy learning loss surface, and alternative ambiguity sets for dsitributionally robust policy learning. More research in any one of these directions would be interesting and I hope this discussion inspires more to tackle these.

\bibliographystyle{agsm}
\bibliography{retargetting_rev2}

\end{document}